\PassOptionsToPackage{hyphens}{url}       % ⇒ url admite cortes de línea en ‘_’
\PassOptionsToPackage{hidelinks}{hyperref}% ⇒ enlaces sin recuadros de color

\documentclass[pdflatex,sn-mathphys-num]{sn-jnl}

\usepackage[T1]{fontenc}
\usepackage[utf8]{inputenc}   % (LuaLaTeX/XeLaTeX no lo necesitan)

\usepackage{graphicx}
\usepackage{multirow}
\usepackage{amsmath,amssymb,amsfonts,amsthm,mathrsfs}
\usepackage[title]{appendix}
\usepackage{xcolor}
\usepackage{textcomp}
\usepackage{manyfoot}
\usepackage{booktabs}
\usepackage{algorithm}
\usepackage{algorithmicx}
\usepackage{algpseudocode}
\usepackage{listings}
\usepackage{float}
\usepackage{subcaption}       % Carga automáticamente caption → NO vuelvas a llamarlo
\usepackage{adjustbox}
\usepackage{makecell}
\usepackage{tabularx}
\usepackage{tikz}
\usepackage{pgfplots}
\usepackage{svg}
\usepackage{setspace}
\usepackage{array}

\setcellgapes{2pt}
\makegapedcells

\theoremstyle{thmstyleone}%
\theoremstyle{thmstyletwo}%

\theoremstyle{thmstylethree}%

\begin{document}

\title[Clinically enhanced explainability for BCC AI]{MultiTask Learning AI system to assist BCC diagnosis with dual explanation}

%%=============================================================%%
%%  AUTORES                                                    %%
%%=============================================================%%

\author*[1]{\fnm{Iván} \sur{Matas}}\email{imatas@us.es}   % añade tu correo
\author[1]{\fnm{Carmen} \sur{Serrano}}\email{cserrano@us.es}
\author[2]{\fnm{Francisca} \sur{Silva-Clavería}}\email{fran.silvaclaveria@gmail.com}
\author[2]{\fnm{Amalia} \sur{Serrano}}\email{amaliaserranog@gmail.com}
\author[3]{\fnm{Tomás} \sur{Toledo-Pastrana}}\email{ttoledop@gmail.com}
\author[1]{\fnm{Begoña} \sur{Acha}}\email{bacha@us.es}

%%=============================================================%%
%%  AFILIACIONES                                               %%
%%=============================================================%%

\affil[1]{%
  \orgdiv{Dpto.\ Teoría de la Señal y Comunicaciones, Escuela Técnica Superior de Ingeniería},%
  \orgname{Universidad de Sevilla},%
  \orgaddress{%
    \city{Seville},%
    \state{Andalucía},%
    \country{Spain}}}

\affil[2]{%
  \orgdiv{Servicio de Dermatología},%
  \orgname{Hospital Universitario Virgen Macarena},%
  \orgaddress{%
    \city{Seville},%
    \state{Andalucía},%
    \country{Spain}}}

\affil[3]{%
  \orgdiv{Hospitales Quirón Salud Infanta Luisa y Sagrado Corazón},%
  \orgname{Quirón Salud},%
  \orgaddress{%
    \city{Seville},%
    \state{Andalucía},%
    \country{Spain}}}

% \author*[1,2]{\fnm{First} \sur{Author}}\email{iauthor@gmail.com}



% \affil*[1]{\orgdiv{Department}, \orgname{Organization}, \orgaddress{\street{Street}, \city{City}, \postcode{100190}, \state{State}, \country{Country}}}

% \affil[2]{\orgdiv{Department}, \orgname{Organization}, \orgaddress{\street{Street}, \city{City}, \postcode{10587}, \state{State}, \country{Country}}}

% \affil[3]{\orgdiv{Department}, \orgname{Organization}, \orgaddress{\street{Street}, \city{City}, \postcode{610101}, \state{State}, \country{Country}}}

%%=============================================================%%
%%  NOTAS OPCIONALES                                           %%
%%=============================================================%%
% \equalcont{Estos autores contribuyeron por igual a este trabajo.}
% \funding{Declaración de financiación, si procede.}
% \conflictsofinterest{Los autores declaran no tener conflictos de interés.}
% \ethics{Aprobaciones éticas correspondientes.}


\abstract{
\textbf{Purpose:} Basal cell carcinoma (BCC) accounts for 75\% of all skin cancers. Currently, all major public hospitals in Spain have a dermatology care protocol that includes teledermatology. This has created an overload for hospital dermatologists, which could be alleviated with an AI tool for prioritization. Several AI systems have been proposed for this purpose, but the lack of transparent diagnostic explanations limits their clinical acceptance and implementation. This fact motivates the present study, which aims to develop an AI tool focused on detecting BCC from dermoscopic images incorporating dermatologist diagnostic criteria to enhance reliability. Specifically, the BCC diagnostic criterium is that a lesion is not considered BCC if it exhibits pigment network pattern, and that a lesion is considered BCC if it exhibits at least one of these BCC patterns: ulceration, ovoid nest, multi globules, maple-leaf, spoke wheel, arborizing telangiectasia. 

\textbf{Methods:} 
We analyzed 1,559 dermoscopic images collected from 60 primary care centers in Andalusia. Four dermatologists annotated the images as exhibiting or not each of the seven possible BCC patterns.
As there is no established Ground Truth to determine the BCC patterns present in a lesion, we propose an Expectation-Maximization consensus algorithm to consolidate the multi-rater annotations into a unified standard reference (SR). 
As an additional novelty, the system incorporates the symbolic reasoning of dermatologists, who base their diagnoses on BCC patterns shown in lesions. To this end, a multitask learning (MTL) system based on MobileNet-V2 was designed. This system can rapidly triage BCC and non-BCC lesions while providing clinical information justifying this classification. This system also provides GradCAM-based maps to dermatologists to improve its reliability and confidence.

\textbf{Results:} 
Three evaluations were performed on the AI system. First, a performance analysis was conducted to evaluate the AI tool's ability to classify lesions as BCC or non-BCC. In this analysis, the model achieved 90\% accuracy (precision=0.90, recall=0.89). The second evaluation analyzed whether the detected patterns agreed with dermoscopic criteria. Notably, at least one clinically relevant BCC pattern was correctly identified in 99\% of BCC-positive cases, and the pigment-network negative criterion was met in 95\% of non-BCC cases. A comparison of the GRADCAM maps with the dermatologist's manual delineation demonstrated strong colocalization with dermatologist-segmented regions (mean foreground density 0.57 vs. background 0.16), confirming alignment of the visual focus of experts.

\textbf{Conclusions:} This work introduces the first clinically validated dual-explanation AI system that combines high-accuracy BCC detection with transparent, pattern-based explanations. This approach closes the critical gap between AI performance and clinical trust in teledermatology, positioning the system for immediate deployment in primary care. Future work will focus on determining the extent to which this dual explanation system improves dermatologists' confidence.

}




\maketitle

\section{Introduction} \label{sec:Introduction}
Worldwide, doctors diagnose skin cancer most commonly, with Melanoma, Basal Cell Carcinoma (BCC), and Squamous Cell Carcinoma (SCC) as the most prevalent types. BCC accounts for approximately $75$\% of all skin cancers, making it the most frequent form. It has well-established clinical criteria for its diagnosis, yet there is significant variability in the presence of these clinical criteria between cases \cite{aad2024,skincancerfoundation2024,cikazynska2018trends,peris2019diagnosis_BCCPatterns}.

The development of public databases has significantly increased the number of published articles related to automatic detection of skin diseases in recent years \cite{Codella2017,combalia2019bcn20000,HAM10000DataBase,isicarchive2024}. Although these databases are accessible and comprehensive for everyone, the clinical criteria used for diagnosis are not readily available. To develop a tool that is useful from a medical perspective, it is crucial to provide not only classification metrics, but also a detailed diagnosis explaining the detected clinical features, offering a more comprehensive result. In this context, Serrano et al. \cite{serrano2022} created a clinically inspired skin lesion classification tool by detecting dermoscopic criteria for BCC.

Current meta-analyses indicate that while AI algorithms significantly outperform general practitioners in skin cancer classification, achieving a sensitivity of 92.5\% versus 64.6\% for generalists, their performance remains comparable to that of expert dermatologists, suggesting that the optimal clinical role for AI is as a collaborative support tool rather than a standalone replacement \cite{salinas2024}. To facilitate this integration, recent research has expanded the scope of AI beyond simple lesion classification. For instance, Liu et al. \cite{liu2024} demonstrated the utility of AI in early screening by predicting skin cancer risk directly from 2D facial images, identifying complex non-linear facial endophenotypes that outperformed traditional risk factor assessments (c-index 0.72 vs. 0.59). Simultaneously, to address the ``black box'' limitations in lesion diagnosis, studies have begun to explicitly encode clinical reasoning into neural architectures. Wang et al. \cite{wang2025} introduced a framework utilizing Clinical Knowledge Graphs to model the directed relationships between dermoscopic attributes of the 7-point checklist, thereby mimicking the structured decision-making process of a clinician.

XAI techniques such as Grad-CAM \cite{Selvaraju_2019_GradCAM} play a crucial role in improving the transparency and interpretability of CNNs in medical diagnosis. By offering visual explanations of the decision-making process, these methodologies allow medical professionals to gain insight into the predictive mechanisms of AI models, thus fostering trust and acceptance in AI-driven diagnoses.

Recent works in dermatological AI have employed various XAI approaches: attention-based mechanisms to highlight diagnostically relevant features \cite{Barata_RNN_XAI}, multimodal explanations combining text and visual information to increase clinician confidence \cite{Chanda2024}, and gradient-based visualization methods like Grad-CAM++ to provide post-hoc interpretability and facilitate detection of misdiagnoses \cite{Rezk2023}. Van der Valden et al. \cite{survey_XAI} provide a comprehensive survey of XAI techniques applied to skin cancer diagnosis. However, these works primarily focus on the explainability of the regions where models focus their attention, providing a post-hoc interpretation of the features extracted by a trained classifier. Thus, clinician trust increases when AI activation overlaps with clinically relevant regions.

In contrast, the key contribution of this work lies in a Multi-Task Learning (MTL) framework that integrates clinical reasoning directly into the model architecture. This dual-layer structure not only performs binary classification of BCC versus non-BCC lesions but also provides pattern-level interpretability grounded in dermoscopic features. By jointly learning these tasks, the system enables a clinically meaningful form of XAI, allowing dermatologists to validate the prediction rationale through pattern-specific attention analysis—thus enhancing the reliability, transparency, and usability of the tool in telemedicine workflows. The practical consequence of integrating clinical reasoning into the model architecture is that the proposed XAI model offers two clinical explanations. First, it provides an explanation of the BCC patterns encountered in the image. Second, it provides a GRADCAM map correlated with the location of these BCC patterns.

This research is driven by the need to improve clinical workflows in healthcare systems based on teledermatology. In modern healthcare, primary care physicians use teledermatology to receive high-quality diagnostic images remotely, allowing preliminary diagnoses of Basal Cell Carcinoma (BCC) using established patterns \cite{peris2019diagnosis_BCCPatterns,peris2002interobserver_BCCPatterns3}. Teledermatology is crucial in areas with limited specialist access and its integration with AI tools promises to enhance diagnostic accuracy and efficiency. Suspected BCC cases are immediately referred to dermatology specialists, improving healthcare efficiency, reducing waiting times, and facilitating early intervention. This work aims to develop an AI tool to assist in this process by providing a binary classification of BCC/non-BCC with interpretable results.

\section{Methodology} \label{sec:Methodology}

\begin{figure}[htpt]
    \centering
    \includegraphics[width=0.8\linewidth]{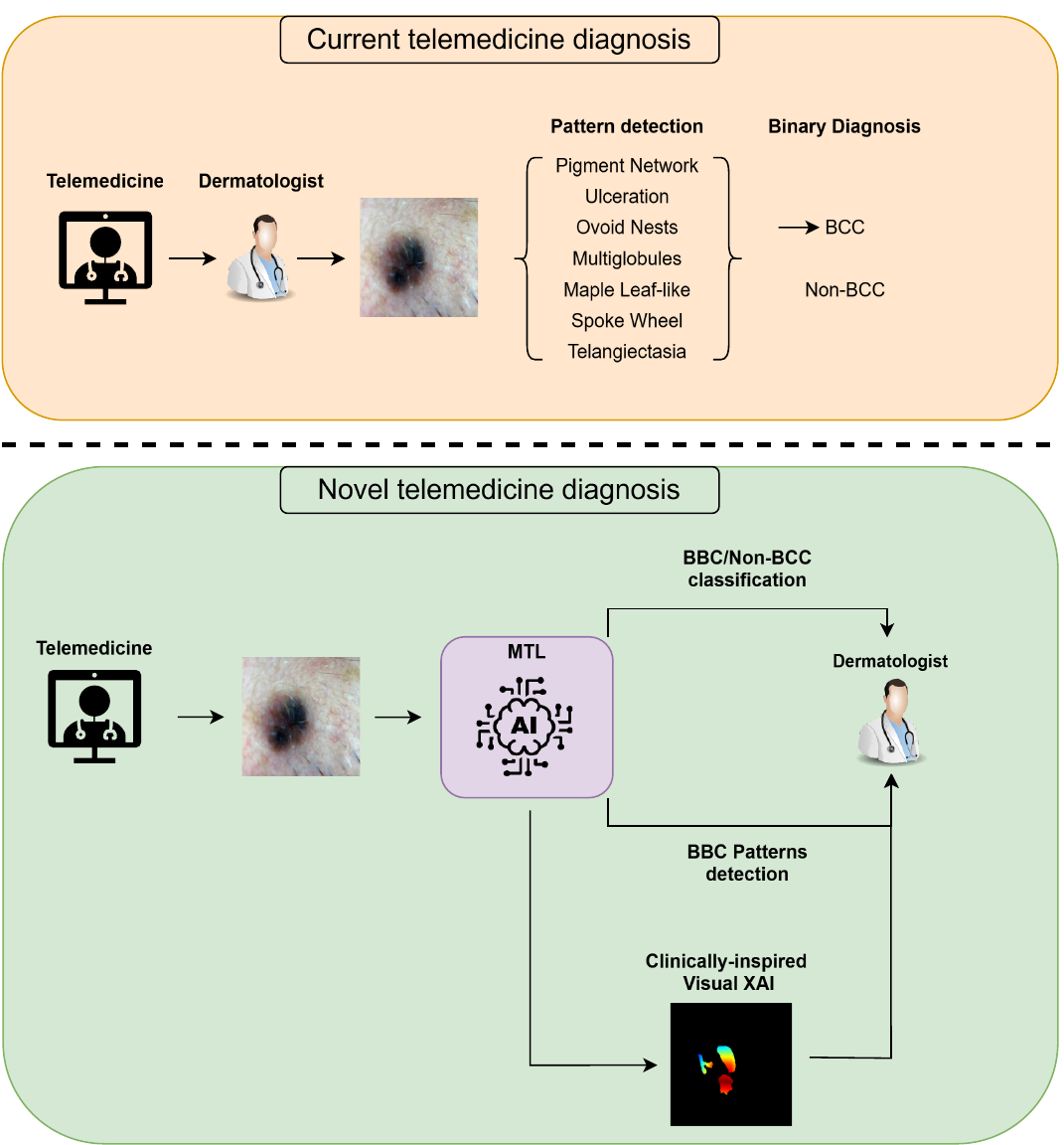}
    \caption{
Comparison between the current and proposed workflows for basal cell carcinoma (BCC) diagnosis via teledermatology. The upper panel (orange) illustrates the standard approach, in which the dermatologist performs manual pattern recognition and binary diagnosis. The lower panel (green) depicts the proposed multi-task learning (MTL) system, where a shared AI backbone simultaneously performs BCC/non-BCC classification and BCC pattern detection, supported by clinically-inspired visual XAI. This novel framework aims to increase automation, interpretability, and clinical alignment in remote diagnosis.
}
    \label{fig:Workflow_Squeme}
\end{figure}

The proposed methodology to improve the teledermatology workflow is summarized in
Fig.~\ref{fig:Workflow_Squeme}, which illustrates both the current and proposed diagnostic workflows. In the proposed workflow, the AI tool not only delivers a BCC/non-BCC diagnosis but also provides a layer of interpretability, guiding the dermatologist to focus on the dermoscopic features most relevant to basal cell carcinoma. This dual-layer explanation ensures that the specialist can validate the AI's predictions by confirming that the model is emphasising the correct clinical patterns, thus reducing the chances of diagnostic errors and increasing trust in AI-assisted teledermatology.

\subsection{Database}\label{sec:database}
The Dermatology Unit of the ``Hospital Universitario Virgen Macarena'' provided the entire database, which consists of dermoscopic images sent from 60 primary care centers during the period 2022--2023. The majority of images used in this study were collected between July and November 2023, coinciding with the deployment of the teledermatology system across these primary care centers. The dataset comprises 1559 dermoscopic images divided into 3 subsets. Four dermatologists provided different types of annotation according to the subsets. Specifically:

\begin{itemize}
    \item The first subset consists of 1089 dermoscopic images that were annotated following a multilabel codification explained in Sect.~\ref{sec:Label_Codification}. To consolidate the multi-rater labeling, an Expectation-Maximization~(EM) algorithm was employed, generating an inferred Standard Reference~(SR), which will be explained in Sect.~\ref{sec:SR_EM}. Subsequently, a binary classification scheme was established to classify into BCC and non-BCC lesions.

    \item An additional set of 334 images was collected, enriched with detailed dermatological annotations beyond the multilabel and binary SR. It includes specialist's manual segmentations of the BCC clinical patterns present in a lesion. Several segmented area may appear on an image if there are some patterns in the BCC lesion. In the Fig.~\ref{fig:Example_Segmentation} an example is shown. These images are designated for the XAI part of the study, enabling a deeper analysis of the model decision process.

    \item The final subset was gathered from the ISIC Archive \cite{isicarchive2024}, utilizing 136 non-BCC images, predominantly consisting of nevus lesions.
\end{itemize}

Images were acquired using different dermoscopic devices deployed across the primary care network, including portable polarized digital dermatoscopes (DermLite II Multispectral, 3Gen) with smartphone adapters \cite{AGSSurSevilla2025, hospitalmacarena2019}, and professional videodermatoscopy systems (FotoFinder Medicam 1000 with D-Scope III) at hospital level \cite{aetsa2006telederma}. This device heterogeneity reflects real-world teledermatology conditions. Regarding image preprocessing, no specific image processing was applied; only images with insufficient quality due to motion blur, poor focus, or inadequate exposure were excluded. The inclusion criterion was adequate visualization of the lesion for clinical assessment. Patient demographic data (age, sex, skin type) were not available due to anonymization protocols required by the ethics committee.

\subsubsection{Label codification}\label{sec:Label_Codification}

\begin{figure}
    \centering
    \includegraphics[width =.8\textwidth]{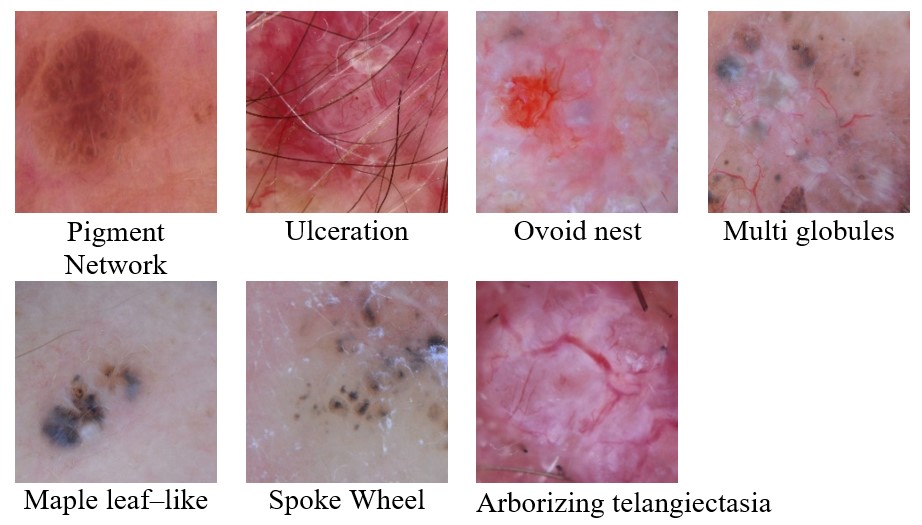}
    \caption{BCC dermoscopic criteria and Pigment Network as a negative criterion.}
    \label{fig:BCC_Patterns}
\end{figure}

Each image may contain multiple dermoscopic patterns. Therefore, we used a one-hot coding scheme to encode the labels during image annotation and subsequently process the dermatologists' annotations. Each image label is a binary word and each BCC dermoscopic pattern is a digit, where $1$ means presence and $0$ means absence. The seven patterns (Fig.~\ref{fig:BCC_Patterns}) that can appear in a BCC lesion are \cite{peris2019diagnosis_BCCPatterns}: Pigment Network~(PN) (negative criterion), Ulceration~(U), Ovoid Nests~(ON), Multiglobules~(MG), Maple Leaf-like~(ML), Spoke Wheel~(SW), Arborizing Telangiectasia~(AT). Thus, each label is a vector of dimensions $[1 \textrm{x} 7]$ using a binary codification where a value of 1 means BCC lesion and a value of 0 means pattern absence. In Table~\ref{tab:CodExamples} there are some examples of this process. In Examples 1 and 2, the labels indicate the presence of BCC dermoscopic patterns. According to clinical criteria, this corresponds to the presence of BCC, which is encoded with the binary label 1. Example 3 indicates the presence of a pigment network, while example 4 indicates that no BCC pattern can be found in the lesion. Thus, the binary encoding must be 0 in both cases, indicating an absence of BCC.

\begin{table}[ht]
\centering
\caption{Example of multilabel and binary encoding for BCC diagnosis}
\label{tab:CodExamples}
\small
\begin{tabular}{l l c l}
\toprule
\textbf{Codification} & \textbf{Multi-label} & \textbf{Binary} & \textbf{Diagnosis} \\
\midrule
Example 1 & [0 1 0 1 1 0 1] & 1 & Presence of BCC \\
Example 2 & [0 0 1 0 0 0 0] & 1 & Presence of BCC \\
Example 3 & [1 0 0 0 0 0 0] & 0 & Absence of BCC \\
Example 4 & [0 0 0 0 0 0 0] & 0 & Absence of BCC \\
\bottomrule
\end{tabular}

\end{table}

\subsection{Standard Reference (SR) Inferring}\label{sec:SR_EM}

The accepted ground truth (GT) for BCC diagnosis is biopsy. However, there is no established GT for BCC dermoscopic patterns, which are subjectively assessed by dermatologists. Several studies have reported a low kappa coefficient when measuring inter-dermatologist agreement in determining the different dermoscopic patterns present in a lesion \cite{polesie2021interobserver, peris2002interobserver_BCCPatterns3}.
Therefore, an adequate SR inferred from the consensus of several dermatologists is required. We implemented an Expectation-Maximization (EM) based algorithm~\cite{dawid1979ExpectationMaximization} to derive a single SR for model training from multiple specialist labels. This algorithm consolidates multilabel annotations from different dermatologists and generates a single inferred SR that encapsulates the collective expertise of the raters. Silva et al. \cite{Invent} used this algorithm to infer the SR from BCC pattern annotations and demonstrated that integrating this diverse expertise mitigates the subjectivity inherent in diagnosing the BCC pattern and improves the reliability and robustness of the classification model.

\begin{table}[ht]
    \centering
    \caption{Sample distribution for binary and multilabel codification.}
    \label{table:Sample_distribution}
    \small
    \renewcommand{\arraystretch}{1.4}
    \begin{tabularx}{\textwidth}{>{\centering\arraybackslash}X >{\centering\arraybackslash}p{1.4cm} *{7}{>{\centering\arraybackslash}X}}
        \toprule
        \multicolumn{2}{c}{\textbf{Binary codification}} & \multicolumn{7}{c}{\textbf{Multi-label codification}} \\
        \midrule
BCC & Non-BCC & PN & U & ON & MG & ML & SW & AT \\
        \midrule
775 & 784 & 557 & 385 & 333 & 191 & 244 & 178 & 455 \\
        \bottomrule
    \end{tabularx}
\end{table}

Table~\ref{table:Sample_distribution} summarises the distribution of labels in the database. As can be seen in this table, the database has a significant class imbalance, with SW and MG patterns under-represented. Several techniques have been used to address this problem, see Sect. \ref{sec:Implementation}.

\subsection{Multi-task learning (MTL) System: BCC Diagnosis with XAI-guided Pattern Recognition}
This section refers to the AI model positioned at the bottom of the Novel telemedicine diagnosis framework in Fig.~\ref{fig:Workflow_Squeme} (Pattern detection AI). The system was developed under a multi-task learning (MTL) paradigm, in which a single backbone jointly supports two interrelated objectives: the binary classification of BCC versus non-BCC lesions, and the detection of clinically relevant BCC dermoscopic patterns that guide this decision. From a clinical point of view, achieving 100\% accuracy in the detection of BCC patterns is not required for a reliable diagnosis: a dermatologist only needs to identify one indicative pattern to confirm the presence of BCC. Accordingly, a clinically aligned XAI system should follow the same principle. The proposed model adheres to this logic: if no pattern is detected, it predicts no BCC; if the PN pattern is detected, it acts as a negative criterion; and if any of the other BCC patterns are detected, the system issues a positive prediction for BCC.
The MTL architecture is based on the MobileNet-V2 model with three classifier layers. The model was coupled with a three-step optimization and training strategy. The first two steps focused on the detection of BCC/non-BCC, while the third step focused on pattern classification. In the first step, ImageNet transfer learning was applied and classifier weights were trained. In the second step, fine-tuning was applied to the last three blocks of the feature extractor, and the classifier was tuned with a lower learning rate (LR) and an increased number of epochs. In the third step, once the column extracts features from the same region as the specialist in the binary task, the classifier was retrained for pattern detection with a very low LR.

In light of the considerable class imbalance evident in Table \ref{table:Sample_distribution}, the methodology was devised to circumvent a direct transition from the adjusted ImageNet weights to patterns characterized by such an imbalance. Instead, an intermediate phase was incorporated, during which the weights were fine-tuned on the dermoscopic images. This intermediate phase allowed the extraction of features specific to dermoscopic images, thereby enhancing the subsequent classification process and ensuring greater accuracy and manageability.

\subsection{Implementation details}\label{sec:Implementation}
Optimising the model was a key component of this work, with a particular focus on enhancing both classification performance and clinical interpretability. To this end, several critical hyperparameters were tuned through empirical experimentation. The training was conducted in two stages. In the first stage, transfer learning from ImageNet was employed using the AdamW schedule-free optimiser \cite{defazio2024road_schedulefree}, with a learning rate of $1\times10^{-5}$ over 100 epochs and a mini-batch size of 32. A dropout rate of 0.3 was selected to regularise the training and reduce the risk of overfitting.

As demonstrated in Sect.~\ref{sec:database}, class imbalance is a significant issue, especially in the multi-label classification of BCC patterns such as ML, MG, or SW. To address this problem, the Focal loss \cite{lin2018focal_loss} was implemented, which is particularly useful when the distribution of samples is not uniform.

Due to the restricted scope of our database, we have employed a stratified \mbox{k-fold} \mbox{cross-validation} to guarantee a thorough assessment of the model's efficacy. Specifically, a $5$-fold stratification \cite{hastie2009elements,Stone-KFold} was used, ensuring that each fold was well-representative of the whole dataset, maintaining the proportion of each class in every fold. This approach mitigates the risk of biased training and testing distributions, which is crucial in datasets with imbalanced class distributions.

For the classification model, we chose MobileNet-V$2$, a lightweight deep neural network known for its efficiency and effectiveness in image classification tasks. To enhance the robustness of the model and account for variabilities in real-world dermoscopic imaging, several data augmentation techniques were applied \cite{AlexNet,cireşan2012multicolumn, simard2003best}. These included rotation to mimic variations in lesion orientation across clinical images, random perspective to simulate different acquisition angles during examination, and Gaussian blur to reflect common issues such as slight motion blur or focus inaccuracies in dermoscopic imaging. These augmentation strategies not only help prevent overfitting, but also enhance the model's ability to generalise to new, unseen images by simulating a variety of imaging conditions in real-world scenarios.

\section{Results}\label{sec:Resutls}

\subsection{Pattern-Based Clinical Reasoning}\label{sec:XAI_Results} %cambiado para no repetir nombre -- Sec2.3
This section analyses the performance of the AI tool for BCC detection in conjunction with the labels provided to explain this classification. Table~\ref{tab:Metrics} presents metrics that summarise this performance. The metrics are averaged over all folds. This table has three parts. The first part shows the performance of the AI tool in binary classification. The second part shows its performance in detecting BCC dermoscopic patterns. Finally, the third part represents the accuracy of the labels that provide the clinical explanation.

In general, the diagnostic performance of BCC / non-BCC is high, around $0.9$ for all metrics. However, the BCC pattern detection performance needs to be analysed with a deeper insight. Minority classes tend to achieve low recall because the AI tool trained with unbalanced databases tends to favour majority classes. As shown in Sect.~\ref{sec:database}, SW, MG and ML are underrepresented classes. Strategies such as data augmentation and advanced sampling, a one-vs-all strategy combined with stratified k-fold cross-validation helped to achieve a more balanced classification across patterns, thereby improving overall model performance. However, the metrics achieved should not be analysed in the same way as BCC/non-BCC performance. They should only be evaluated to the extent that they provide a correct explanation for binary classification. For the specialist, a tool that perfectly classifies all BCC patterns within a lesion is not particularly useful, as identifying even a single pattern can already indicate a positive BCC diagnosis. Therefore, the third section of Table \ref{tab:Metrics} provides information relevant to the clinician by assessing the model's ability to detect at least one of the BCC patterns annotated by specialists, which is sufficient for a positive BCC diagnosis according to clinical criteria. Furthermore, this information is further supported by Table \ref{tab:Mean_Values}, which, as will be discussed later, shows whether the classification of each pattern is being extracted from the same region as identified by the dermatologist. This additional evaluation is summarised in the third section of Table \ref{tab:Metrics}, titled ``Clinically-inspired XAI''. The table demonstrates that 73\% of non-BCC lesions without any BCC patterns, 95\% of non-BCC lesions with PN, and 99\% of BCC lesions with at least one BCC pattern are correctly labelled as such.

\begin{table}[ht]
    \centering
    \caption{Evaluation using binary, multilabel classification metrics and physician-guided analysis.}
    \label{tab:Metrics}
    \small
    \begin{tabularx}{\textwidth}{l *{4}{>{\centering\arraybackslash}X}}
        \toprule
& \makecell{Recall \\ ($\sigma^{2}$)} & \makecell{Specificity \\ ($\sigma^{2}$)} & \makecell{Precision \\ ($\sigma^{2}$)} & \makecell{Accuracy \\ ($\sigma^{2}$)} \\
        \midrule
        \textbf{BCC/Non-BCC} & & & & \\
        \midrule
& 0.89 & 0.89 & 0.90 & 0.90 \\
        \midrule
        \textbf{Pattern detection} & & & & \\
        \midrule
Pigment Network & 0.94 ($4.19 \cdot 10^{-4}$) & 0.96 ($1.07 \cdot 10^{-4}$) & 0.97 ($2.47 \cdot 10^{-4}$) & 0.95 ($4.50 \cdot 10^{-6}$) \\
Ulceration & 0.81 ($4.14 \cdot 10^{-3}$) & 0.75 ($8.94 \cdot 10^{-4}$) & 0.52 ($8.94 \cdot 10^{-4}$) & 0.77 ($4.84 \cdot 10^{-4}$) \\
Ovoid Nests & 0.65 ($2.48 \cdot 10^{-3}$) & 0.84 ($1.52 \cdot 10^{-4}$) & 0.53 ($3.78 \cdot 10^{-5}$) & 0.84 ($1.20 \cdot 10^{-5}$) \\
Multiglobules & 0.61 ($6.16 \cdot 10^{-3}$) & 0.81 ($2.05 \cdot 10^{-3}$) & 0.32 ($2.15 \cdot 10^{-3}$) & 0.80 ($1.10 \cdot 10^{-3}$) \\
Maple Leaf-like & 0.50 ($8.76 \cdot 10^{-3}$) & 0.82 ($9.43 \cdot 10^{-5}$) & 0.34 ($1.79 \cdot 10^{-3}$) & 0.77 ($2.47 \cdot 10^{-4}$) \\
Spoke Wheel & 0.60 ($2.63 \cdot 10^{-3}$) & 0.87 ($8.09 \cdot 10^{-4}$) & 0.37 ($1.15 \cdot 10^{-3}$) & 0.84 ($3.93 \cdot 10^{-4}$) \\
Arborizing Telangiectasia & 0.89 ($1.08 \cdot 10^{-3}$) & 0.76 ($4.57 \cdot 10^{-4}$) & 0.61 ($4.69 \cdot 10^{-4}$) & 0.80 ($2.83 \cdot 10^{-4}$) \\
        \midrule
        \textbf{Clinically-inspired XAI} & & & & \\
        \midrule
No-pattern (all zeros) & -- & -- & -- & 0.73 \\
Pigment Network & 0.94 & 0.96 & 0.97 & 0.95 \\
BCC pattern detection & 0.84 & 0.88 & 0.71 & 0.99 \\
        \bottomrule
    \end{tabularx}
\end{table}

\subsection{Clinically-inspired Visual XAI}\label{sec:Visual_XAI_Results}
\subsubsection{Clinical Validation Framework}
This section refers to the AI model at the top of Novel telemedicine diagnosis in Fig.~\ref{fig:Workflow_Squeme} (BCC/Non-BCC classification). From a clinical perspective, the utility of an XAI system extends beyond simple visualization of model attention—it must demonstrate that the model's focus aligns with the diagnostic criteria used by dermatologists. To address this requirement, we quantify the accuracy of the AI tool in focusing on the correct part of the lesion, specifically the BCC dermoscopic patterns identified by clinicians. To this end, the areas of the BCC pattern delineated by dermatologists are compared with the activated areas of the model, providing a quantitative measure of the model's agreement with human diagnostic criteria and demonstrating its ability to accurately identify critical features of BCC lesions.

To develop this concept, expert-generated segmentations are used. As detailed in Sect.~\ref{sec:database}, a subset of 334 samples was segmented by a specialist. A dermatologist performed an individual segmentation of each BCC pattern for each image (Fig.~\ref{fig:Example_Segmentation}). These expert annotations serve as the clinical ground truth against which the model's attention mechanisms are validated, ensuring that XAI methods are not only technically robust but also tailored to clinical reasoning processes.

\subsubsection{Expert-Model Agreement Assessment}
To complement visual inspection, we quantify the spatial agreement between Grad-CAM activations and expert segmentations using conditional probability density functions, as described below.

To quantify the accuracy of the model activation areas with respect to the areas of clinical interest, the conditional probability density functions of the normalised Grad-CAM values within and outside the area segmented by dermatologist were estimated. Let $z(x,y)$ the Grad-CAM value at position~$(x,y)$. Denote Fg the area segmented by the dermatologist and Bg the background. $P\left(z\left(x,y\right) \mid \text{Fg}\right)$ is the probability density function of the Grad-CAM values for pixels $(x,y) \in \text{Fg}$ and $P\left(z\left(x,y\right) \mid \text{Bg}\right)$ is the probability density function of Grad-CAM values for pixels $(x,y) \in \text{Bg}$.

\begin{figure}[htp]
    \centering
    \resizebox{\textwidth}{!}{%
        \begin{minipage}{\textwidth}
            \centering
            \begin{subfigure}{0.3\textwidth}
                \centering
                \includegraphics[width=\textwidth]{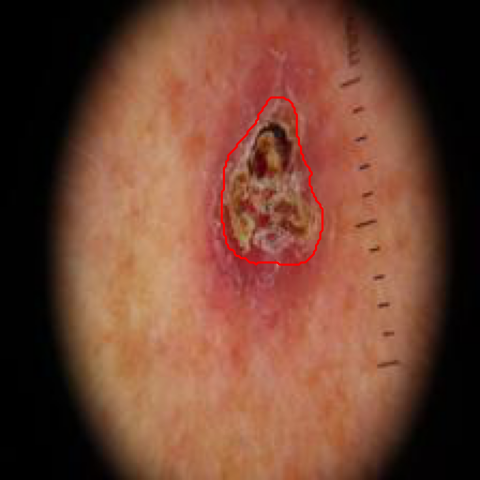}
                \caption{Original image and the detailed BCC pattern segmentation by a specialist in red.}
                \label{fig:Segmentation_083939_445997}
            \end{subfigure}
            \hfill
            \begin{subfigure}{0.3\textwidth}
                \centering
                \includegraphics[width=\textwidth]{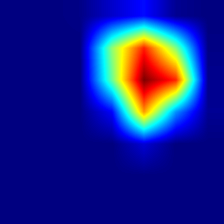}
                \caption{GradCAM heatmap showcasing areas of highest model activation.}
                \label{fig:GradCAMHeatmap_083939_445997}
            \end{subfigure}
            \hfill
            \begin{subfigure}{0.3\textwidth}
                \centering
                \includegraphics[width=\textwidth]{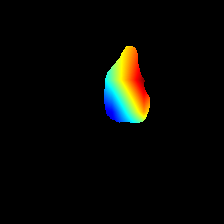}
                \caption{Integration of Grad-CAM heatmap and specialist segmentation.}
                \label{fig:SegGradCAM_083939_445997}
            \end{subfigure}
            \begin{subfigure}{\textwidth}
                \centering
                \includegraphics[width=0.7\textwidth]{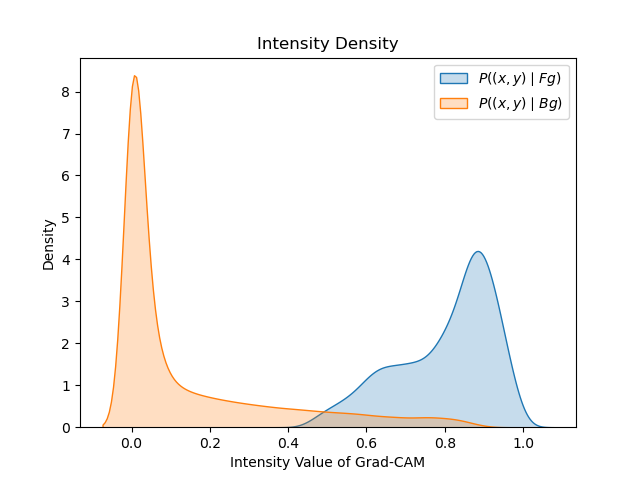}
                \caption{Analysis of Grad-CAM heatmap and expert segmentation. The blue density curve indicates that a significant portion of the Grad-CAM is accurately segmented by the expert.}
                \label{fig:GradCAMSegmentationProcessing_083939_445997}
            \end{subfigure}
        \end{minipage}%
}
    \caption{ Multi-panel visualization demonstrating the clinical validation of Grad-CAM activations against expert segmentations. (a) displays the original dermoscopic image with the expert segmentation boundary (red contour) delineating the clinically relevant lesion region; (b) shows the corresponding Grad-CAM heatmap indicating regions of highest model activation (intensity scale from low to high); (c) overlays the dermatologist's manual segmentation on the Grad-CAM heatmap to assess spatial alignment; panel (d) presents the conditional probability density functions comparing Grad-CAM activation distributions inside (blue curve) versus outside (orange curve) the clinically relevant regions. The clear separation between distributions demonstrates that the model's attention is appropriately focused on dermoscopically significant areas identified by clinical experts. This example illustrates a correctly predicted BCC case where the model's decision-making aligns with clinical diagnostic criteria.}
    \label{fig:GradCAMSegmentation_083939_445997}
\end{figure}

Fig.~\ref{fig:GradCAMSegmentation_083939_445997} illustrates this analysis. Fig.~\ref{fig:Segmentation_083939_445997} shows the original BCC lesion. Fig.~\ref{fig:GradCAMHeatmap_083939_445997} shows the Grad-CAM map. Fig.~\ref{fig:SegGradCAM_083939_445997} shows the dermatologist's segmentation overlaid on the Grad-CAM map. Fig.~\ref{fig:GradCAMSegmentationProcessing_083939_445997} shows an example of the two conditional probability density functions. The orange curve represents $P\left(z\left(x,y\right) \mid \text{Bg}\right)$, and the blue curve represents $P\left(z\left(x,y\right) \mid \text{Fg}\right)$. The orange curve is centred near 0, indicating low activation outside the mask, while the blue curve shows significant Grad-CAM information within the clinical segmentation, indicating that the model extracts features from the same region as the specialist.

Table~\ref{tab:Mean_Values} summarises the information extracted from these probability density functions. Specifically, the mean, standard deviation of $z\left(x,y\right)$ for $(x,y) \in Fg$ and $(x,y) \in \text{Bg}$ respectively, and the intersection area between $P\left(z\left(x,y\right) \mid \text{Fg}\right)$ and $P\left(z\left(x,y\right) \mid \text{Bg}\right)$ are shown. This table highlights that correctly predicted samples exhibit a higher average $P\left(z\left(x,y\right) \mid \text{Fg}\right)$ value compared to misclassified samples. In addition, they display a lower average $P\left(z\left(x,y\right) \mid \text{Bg}\right)$ value, which suggests that the model is focusing on regions akin to those a dermatologist would examine when diagnosing the lesion.

To complement the descriptive analysis of Grad-CAM conditional distributions for correctly and incorrectly classified cases (upper part of Table~\ref{tab:Mean_Values}), we performed an image-wise statistical comparison between activations inside and outside the dermatologists' masks in the energy domain $z^2(x,y)$. As reported in the lower part of Table~\ref{tab:Mean_Values}, statistically significant differences between foreground and background energy were observed in 328/334 images (98.2\,\%) at $\alpha = 0.05$, with 322/334 images (96.5\,\%) remaining significant after Bonferroni correction and 328/334 images (98.2\,\%) after FDR (Benjamini--Hochberg) correction. The corresponding mean Cohen's $d$ of 1.972 indicates a very large effect size, confirming that Grad-CAM concentrates substantially more energy within the region of clinical interest than in the surrounding background and providing strong statistical evidence for the foreground--background activation contrast.

\begin{figure}[htp]
    \centering
    \resizebox{0.9\textwidth}{!}{%
    \begin{minipage}{1\textwidth}
        \begin{subfigure}[t]{0.3\textwidth}
            \centering
            \includegraphics[width=\textwidth]{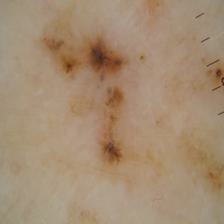}
            \caption{Image example of specialist's segmentation.}
            \label{fig:first}
        \end{subfigure}
        \hfill
        \begin{subfigure}[t]{0.3\textwidth}
            \centering
            \includegraphics[width=\textwidth]{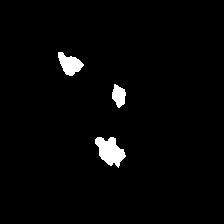}
            \caption{Full lesion segmentation with all pattern segmentations.}
            \label{fig:second}
        \end{subfigure}
        \hfill
        \begin{subfigure}[t]{0.3\textwidth}
            \centering
            \includegraphics[width=\textwidth]{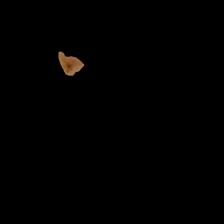}
            \caption{Maple Leaf-like pattern present on the original image.}
            \label{fig:third}
        \end{subfigure}
        \hfill
        \begin{subfigure}[t]{0.3\textwidth}
            \centering
            \includegraphics[width=\textwidth]{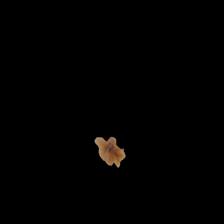}
            \caption{Second Maple Leaf-like pattern present on the original image.}
            \label{fig:fourth}
        \end{subfigure}
        \hfill
        \begin{subfigure}[t]{0.3\textwidth}
            \centering
            \includegraphics[width=\textwidth]{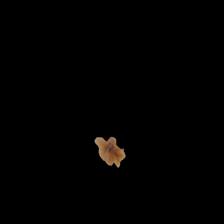}
            \caption{Ovoid Nests pattern present on the original image.}
            \label{fig:five}
        \end{subfigure}
        \hfill
        \begin{subfigure}[t]{0.3\textwidth}
            \centering
            \includegraphics[width=\textwidth]{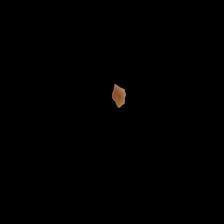}
            \caption{Multiglobules pattern present on the original image.}
            \label{fig:six}
        \end{subfigure}
    \end{minipage}
} % end resizebox
    \caption{ Comprehensive example of the specialist segmentation process for pattern-based clinical validation.  (a) shows the original dermoscopic image with multiple BCC patterns present; (b) displays the complete integrated segmentation combining all clinically relevant patterns identified by the dermatologist; (c)–(f) show individual pattern segmentations for Maple Leaf-like appearance (two instances), Ovoid Nests, and Multiglobules. This systematic segmentation approach enables pixel-level validation of the model's attention mechanisms against clinically defined regions, ensuring that the AI tool's visual explanations align with expert dermatological knowledge. The red contours indicate manually segmented regions of clinical interest used for quantitative comparison with Grad-CAM activation maps.}
    \label{fig:Example_Segmentation}
\end{figure}

The DICE and Jaccard coefficients are commonly used to evaluate accuracy. However, in this study, these metrics were not applicable due to the significant difference in detail between manual segmentations by specialists and the more generalized regions highlighted by Grad-CAM (Fig. \ref{fig:Example_Segmentation}). Activation maps tend to cover larger areas than the precise segmentations, making these coefficients less representative for assessing model performance.

\begin{table}
    \centering
    \caption{Summary of Grad-CAM descriptive statistics and energy-based validation inside and outside the region of clinical interest.The upper section reports conditional means and standard deviations of foreground and background activations.
The lower section presents the statistical comparison of integrated energy ($\sum z^2$) between both regions, including the proportion of significant $z$-tests with multiple-comparison corrections and the corresponding mean effect size (Cohen’s $d$), quantifying the strength of the foreground–background contrast.
}
    \label{tab:Mean_Values}
    \small
    \begin{tabularx}{\textwidth}{>{\centering\arraybackslash}X c *{4}{>{\centering\arraybackslash}X}}
        \toprule
        \multicolumn{6}{c}{\textbf{Descriptive statistics (conditional PDFs)}} \\
        \midrule
        \makecell{Prediction} & \makecell{Intersection} &
        \makecell{Mean \\ $P\left(z(x,y) \mid \text{Fg}\right)$} &
        \makecell{Mean \\ $P\left(z(x,y) \mid \text{Bg}\right)$} &
        \makecell{Std \\ $P\left(z(x,y) \mid \text{Fg}\right)$} &
        \makecell{Std \\ $P\left(z(x,y) \mid \text{Bg}\right)$} \\
        \midrule
Correct & 0.24 & 0.57 & 0.16 & 0.14 & 0.22 \\
Incorrect & 0.32 & 0.33 & 0.14 & 0.01 & 0.21 \\
        \midrule
        \midrule
        \multicolumn{6}{c}{\textbf{Statistical validation (inside vs. outside mask)}} \\
        \midrule
        \makecell{Domain} &
        \makecell{$z$-test \\ (uncorrected)} &
        \makecell{$z$-test \\ Bonferroni} &
        \makecell{$z$-test \\ FDR (BH)} &
        \makecell{Cohen's $d$ \\ (mean)} &
        \makecell{Effect size \\ interpretation} \\
        \midrule
        \makecell{Energy \\ ($z^2$)} &
        \makecell{(98.2\,\%)} &
        \makecell{(96.5\,\%)} &
        \makecell{(98.2\,\%)} &
1.972 &
        \makecell{Very large \\ effect} \\
        \bottomrule
    \end{tabularx}
\end{table}

However, additional information for validation is provided. For this study, the segmentations of individual BCC patterns (see~\ref{sec:database}) were combined into a single segmented image, as shown in Fig.~\ref{fig:Example_Segmentation}. The manual segmentations were then overlayed with activation maps to validate the clinical information provided by Grad-CAM. This not only confirmed whether the network is correctly trained in BCC/non-BCC detection, allowing us to use that spine for pattern detection as previously explained, but also identified the dermoscopic pattern of that region verifying if indeed clinical XAI is working correctly.

The conditional-PDF analysis shows that correctly classified lesions exhibit significantly higher activation inside clinically relevant regions than outside, reinforcing Grad-CAM as a trustworthy, clinically aligned explanation tool for BCC detection.
\subsection{Model Selection and Deployment Considerations}

A critical consideration in developing AI tools for clinical deployment is the balance between model performance and computational efficiency. In the context of teledermatology for primary care settings in Andalusia, where hardware resources are often limited and real-time inference is essential, we selected MobileNet-V2 as the backbone architecture. To validate this architectural choice, we compared its performance against more complex state-of-the-art architectures, specifically ConvNext Tiny and Vision Transformer (ViT-Base B/16), using the same training protocol and dataset.

Table~\ref{tab:ModelComparison} presents a comparative analysis of these architectures. While ConvNext Tiny and ViT-Base contain 8x and 25x more parameters respectively, and require 15x and 59x more computational operations (FLOPs), they do not provide meaningful improvements in pattern detection performance—the most challenging aspect of our task given the limited samples for minority BCC patterns. In fact, MobileNet-V2 achieves comparable or superior pattern detection accuracy (0.82) compared to ConvNext Tiny (0.77) and ViT-Base (0.78), despite being significantly more lightweight.

\begin{table}[ht]
    \centering
    \caption{Comparison of model architectures for BCC pattern detection and computational efficiency.}
    \label{tab:ModelComparison}
    \small
    \begin{tabular}{lcccc}
        \toprule
Model & Parameters & FLOPs & BCC/Non-BCC Acc. & Pattern Avg Acc. \\
        \midrule
MobileNet-V2 (ours) & 3.5M & 0.3G & 0.90 & 0.82 \\
ConvNext Tiny & 28M & 4.5G & 0.90 & 0.77 \\
ViT-Base (B/16) & 86M & 17.6G & 0.86 & 0.78 \\
        \bottomrule
    \end{tabular}
\end{table}

This finding is particularly significant for clinical deployment: MobileNet-V2's compact size (14 MB) and minimal computational requirements enable rapid inference on standard clinical workstations without requiring specialized GPU hardware, making it practical for integration into primary care workflows. The ability to deploy an effective diagnostic aid system without substantial infrastructure investment is essential for widespread adoption in resource-constrained healthcare settings. Our results demonstrate that architectural efficiency and clinical deployability need not come at the cost of diagnostic performance, particularly when the model is designed with domain-specific considerations such as pattern-based reasoning and clinical validation.

\section{Discussion}

The aim of this work was to develop a tool that is truly useful to dermatologists by incorporating clinically relevant information, such as BCC patterns and their segmentation. The tool provides not only a BCC/non-BCC classification but also a deeper level of interpretability through detailed pattern analysis. However, further refinement is needed to ensure that the tool can deliver not only a binary diagnosis of BCC/non-BCC but also a comprehensive identification of specific patterns with accuracy percentages for each. This level of detail aims to make the information truly valuable for teledermatology applications.

The two essential criteria for the clinical diagnosis of BCC by dermatologists are: (1) the presence or absence of any BCC pattern, and (2) the Pigment Network as a negative criterion. The Clinically-inspired XAI tool correctly detects 99\% of cases where any of the six BCC patterns are present in the lesion. Additionally, we aimed to verify that the tool adheres closely to clinical criteria, which is why the double explanation approach was implemented, as illustrated in the results.

As demonstrated in Table \ref{tab:Metrics}, the clinical reliability of the diagnosis was confirmed to be robust for specialists. Furthermore, Table \ref{tab:Mean_Values} indicates that the feature extraction for each dermoscopic pattern was consistently performed from clinically relevant regions, aligning the model's attention with the areas identified by dermatologists. This alignment validates the model's ability to not only distinguish between BCC and non-BCC lesions but also to provide precise pattern-based explanations that improve interpretability and clinical trust.

In the literature, there are other studies that have developed tools aimed at addressing similar challenges. Nevertheless, the tool presented in this work uniquely combines BCC/non-BCC detection with an explainability component for BCC patterns, which has been clinically validated as shown in Table \ref{tab:Mean_Values}. This additional layer of clinical interpretability provides a distinctive advantage that existing tools lack, significantly enhancing its value in a medical context.

Understanding the decision-making process of convolutional neural networks (CNNs) is critical for clinical applications where interpretability is as important as performance metrics. The most common visualization techniques in XAI are activation maps and gradient-weighted class activation mapping (Grad-CAM). Activation Maps, as described by Zhou et al. \cite{zhou2015learning_CAM}, extract features from different network layers to show which patterns the model focuses on, although they may not always directly correlate with specific outcomes. On the other hand, Grad-CAM, as described in detail by Selvaraju et al. \cite{Selvaraju_2019_GradCAM}, uses the gradients of the predicted class that flow into the final convolutional layer to produce a localization map. The clinically-inspired explanation that we proposed in this paper is based on Grad-CAM.

While gradient-based visualization methods such as Grad-CAM \cite{Selvaraju_2019_GradCAM} are widely used in medical AI for post-hoc interpretability, our approach extends beyond standard visual explanations by integrating rigorous clinical validation through expert segmentations and conditional probability density functions. This quantitative validation ensures that the model's attention aligns with diagnostically relevant regions identified by dermatologists, transforming Grad-CAM from a purely visual tool into a clinically validated explanation mechanism. Combined with our pattern-based reasoning layer, this dual-explanation framework provides both what the model detects (specific BCC patterns) and where it focuses (validated against expert knowledge), addressing the critical need for clinically aligned XAI in dermatological diagnosis.

The implementation of this tool will allow the development of a new primary care protocol in Andalusia, streamlined the workflow between primary care and dermatologists. Its deployment would be highly efficient, as the entire system is built using the lightweight and efficient MobileNet model, which will facilitate this implementation. In addition, its strong clinical reliability provides a solid foundation for adoption by specialists, ensuring both accuracy and ease of use in clinical settings.

\section{Conclusions and Implications}

This work presents the first clinically validated dual-explanation AI system for Basal Cell Carcinoma detection that integrates pattern-based clinical reasoning directly into the model architecture. By combining Multi-Task Learning with explainable AI techniques, we bridge the critical gap between AI performance and clinical trust in teledermatology applications.

Our system achieves 90\% accuracy in binary BCC/non-BCC classification while simultaneously providing clinically meaningful explanations through two complementary layers. The pattern-based explanation layer correctly identifies at least one BCC pattern in 99\% of BCC-positive cases and properly recognizes the pigment network negative criterion in 95\% of non-BCC cases. The visual explanation layer, validated through Grad-CAM analysis, demonstrates strong spatial agreement with dermatologist-segmented regions, with mean activation values of 0.57 inside clinically relevant areas compared to 0.16 in background regions. This dual-layer approach ensures that the AI system not only provides accurate diagnoses but also explanations that align with dermatological diagnostic criteria, addressing the interpretability requirements essential for clinical acceptance.

The clinical implications of this work are substantial. The lightweight MobileNet-V2 architecture enables efficient deployment in resource-constrained settings, making it particularly suitable for the teledermatology workflow between primary care centers and hospital dermatology units in Andalusia. By providing both a BCC/non-BCC diagnosis and pattern-level explanations, the system allows dermatologists to validate AI predictions through familiar clinical criteria, potentially reducing diagnostic errors and increasing confidence in AI-assisted decision-making. This approach directly addresses the current overload faced by hospital dermatologists managing cases from 60 primary care centers, offering a tool that not only aids in triage but also enhances the educational value of each case through transparent, clinically grounded explanations.

The integration of clinical reasoning into the model architecture, rather than relying solely on post-hoc explanation methods, represents a significant methodological advancement. Unlike previous approaches that use XAI techniques primarily to visualize regions of model attention, our MTL framework jointly learns classification and pattern detection tasks, enabling the system to reason about BCC in a manner aligned with clinical practice. This design ensures that explanations are not merely illustrative but are intrinsically connected to the diagnostic process itself.

Several avenues for future research emerge from this work. First, conducting reader studies with dermatologists would provide valuable insights into how the dual-explanation system affects diagnostic confidence and decision-making in real clinical scenarios. Second, extending the approach to other skin lesion types and incorporating additional dermoscopic criteria could broaden the system's clinical utility. Third, investigating the integration of this tool into existing teledermatology platforms and evaluating its impact on workflow efficiency and diagnostic outcomes in prospective clinical trials would provide evidence for wider deployment. Finally, exploring the generalizability of the MTL-XAI framework to other medical imaging domains where clinical criteria guide diagnosis could extend the impact of this work beyond dermatology.

In conclusion, this work demonstrates that high-performance AI systems for medical diagnosis can be designed to provide transparent, clinically aligned explanations without sacrificing accuracy. By integrating pattern-based reasoning and visual explanations validated against expert annotations, we have developed a tool positioned for immediate deployment in primary care teledermatology, with the potential to improve both diagnostic efficiency and clinical trust in AI-assisted healthcare.

\clearpage
\bibliography{sn-bibliography}% common bib file
\clearpage

\backmatter

\section*{Statements and Declarations}

\bmhead{Funding}  
The payment for the publication of this paper was covered by the project PID2021-127871OB-I00, funded by the MICIU/AEI/10.13039/501100011033. This work was supported by the Andalusian Regional Government (PROYEXCEL\_00889) and is part of the projects PID2021-127871OB-I00 and PID2024-157491OB-I00, both funded by the MICIU/AEI/10.13039/501100011033 and co-funded by the European Union through the ERDF/European Union (NextGenerationEU/PRTR). 

\bmhead{Competing Interests}  
The authors declare that they have no competing interests to disclose.

\bmhead{Ethical approval and consent}  
This study was conducted in accordance with the Declaration of Helsinki and applicable ethical and legal standards. Ethical approval was obtained from the Comité Coordinador de Ética de la Investigación Biomédica de Andalucía (Spain), which issued a favourable opinion on 28 March 2023 (Acta 3/23), under protocol code PEIBA 1901-N-22.

Written informed consent was obtained from all participants for the use of their dermoscopic images for research purposes. All images and associated metadata were anonymized prior to analysis, and the dataset does not contain any sensitive or personally identifiable information. Data processing complied with the General Data Protection Regulation (Regulation (EU) 2016/679) and the Spanish Organic Law 3/2018 on data protection.

\bmhead{Author Contributions}  
\begin{itemize}
    \item \textbf{Iván Matas}: Writing – Original Draft Preparation, Formal Analysis, Investigation, Methodology, Software, Validation, Visualization, Writing – Review \& Editing
    \item \textbf{Carmen Serrano}: Supervision, Methodology, Validation, Conceptualization, Funding Acquisition, Project Administration
    \item \textbf{Francisca Silva-Clavería}: Data Curation, Resources
    \item \textbf{Amalia Serrano}: Data Curation, Resources
    \item \textbf{Tomás Toledo-Pastrana}: Data Curation, Resources
    \item \textbf{Begoña Acha}:Supervision, Methodology, Validation, Conceptualization, Funding Acquisition, Project Administration
\end{itemize}

\bmhead{Data Availability}  
The dermoscopic images obtained from institutional clinical sources are not publicly available due to legal and ethical restrictions related to personal data protection. In particular, Spanish data protection law (Organic Law 3/2018, of 5 December, on the Protection of Personal Data and guarantee of digital rights) restricts international transfers of pseudonymized and/or anonymized health data outside the scope of application of the General Data Protection Regulation (Regulation (EU) 2016/679).

As a result, these data can only be accessed under controlled conditions and subject to approval by the responsible institution and Ethics Committee. Researchers interested in accessing the data may contact the corresponding author to explore potential access under a formal data use agreement and in compliance with applicable legal and ethical requirements.

This study also includes dermoscopic images that are publicly available through the ISIC 2019 Challenge dataset.

To support reproducibility, a CSV file listing the identifiers of all images used in the study, including those sourced from the ISIC 2019 Challenge, as well as the corresponding dataset splits, will be made publicly available in the project’s GitHub repository \url{https://github.com/Mataas18/BCC_Patterns_XAI}.

\end{document}